%% file: main.tex
\documentclass{article}
\usepackage[margin=1in]{geometry}

\PassOptionsToPackage{numbers}{natbib}

\usepackage[utf8]{inputenc} 
\usepackage[T1]{fontenc}    
\usepackage{hyperref}       
\usepackage{url}            
\usepackage{booktabs}       
\usepackage{amsfonts}       
\usepackage{nicefrac}       
\usepackage{microtype}      
\usepackage{xcolor}         
\usepackage{enumitem}

\usepackage{mathtools}
\usepackage{amsmath}
\usepackage{enumitem}
\usepackage[mode=build]{standalone}
\usepackage{pdfpages}
\usepackage{float}
\usepackage{amsthm}
\usepackage{pgf}
\usepackage{placeins}
\usepackage{makecell}
\usepackage{natbib}

\usepackage{caption}
\usepackage[parfill]{parskip}

\usepackage[capitalize,noabbrev]{cleveref}

\theoremstyle{plain}
\newtheorem{theorem}{Theorem}[section]

\newtheorem{lemma}[theorem]{Lemma}

\theoremstyle{definition}
\newtheorem{definition}[theorem]{Definition}
\newtheorem{assumption}[theorem]{Assumption}
\theoremstyle{remark}
\newtheorem{remark}[theorem]{Remark}
\usepackage{svg}

\usepackage{array}
\newcolumntype{?}{!{\vrule \vrule}}

\newcommand{\indep}{\perp \!\!\!\! \perp}
\newcommand{\notimplies}{%
  \mathrel{{\ooalign{\hidewidth$\not\phantom{=}$\hidewidth\cr$\implies$}}}}

\title{Target Conditioned Representation Independence (TCRI); From Domain-Invariant to Domain-General Representations}

\author{
\textbf{Olawale Salaudeen}\\
Department of Computer Science\\
University of Illinois at Urbana-Champaign\\
\texttt{oes2@illinois.edu}
\and
\textbf{Oluwasanmi Koyejo}\\
Department of Computer Science\\
Stanford University \\
\texttt{sanmi@stanford.edu}
}

\date{}

\begin{document}
\maketitle
\begin{abstract}
 We propose a Target Conditioned Representation Independence (TCRI) objective for domain generalization. TCRI addresses the limitations of existing domain generalization methods due to incomplete constraints.
 Specifically, TCRI implements regularizers motivated by conditional independence constraints that are sufficient to strictly learn complete sets of invariant mechanisms, which we show are necessary and sufficient for domain generalization. Empirically, we show that TCRI is effective on both synthetic and real-world data. TCRI is competitive with baselines in average accuracy while outperforming them in worst-domain accuracy, indicating desired cross-domain stability.
\end{abstract}

\section{Introduction}
Machine learning models are evaluated by their ability to generalize (generate reasonable predictions for unseen examples). Often, learning frameworks are designed to exploit some shared structure between training data and the expected data at deployment. A common assumption is that the training and testing examples are drawn independently and from the same distribution (iid). Given this iid assumption, Empirical Risk Minimization (ERM; \citep{NIPS1991_ff4d5fbb}) gives strong generalization guarantees and is effective in practice.

Nevertheless, many practical problems contain distribution shifts between train and test domains, and ERM can fail under this setting \citep{Arjovsky2019InvariantRM}. This failure mode has impactful real-world implications. For example, in safety-critical settings such as autonomous driving \citep{Amodei2016ConcretePI, Filos2020CanAV}, where a lack of robustness to distribution shift can lead to human casualties; or in ethical settings such as healthcare, where distribution shifts can lead to biases that adversely affect subgroups of the population \citep{singh2021fairness}. Many works have developed approaches for learning under distribution shift to address this limitation. 
Among the various strategies to achieve domain generalization, Invariant Causal Predictions (ICP; \citep{Peters2015CausalIU}) has emerged as popular. ICPs assume that while some aspects of the data distributions may vary across domains, the causal structure (or data-generating mechanisms) remains the same and try to learn those domain-general causal predictors.

Following ICP, \citep{Arjovsky2019InvariantRM} propose Invariant Risk Minimization (IRM) to identify invariant mechanisms by learning a representation of the observed features that yields a shared optimal linear predictor across domains. However, recent work~\citep{Rosenfeld2021TheRO}, has shown that the IRM objective does not necessarily strictly identify the causal predictors, i.e., the representation learn may include noncausal features. Thus, we investigate the conditions necessary to learn the desired domain-general predictor and diagnose that the common domain-invariance Directed Acyclic Graph (DAG) constraint is insufficient to (i) strictly and (ii) wholly identify the set of causal mechanisms from observed domains. This insight motivates us to specify appropriate conditions to learn domain-general models which we propose to implement using regularizers.

{\bf Contributions. } We show that neither a strict subset nor superset of existing invariant causal mechanisms is sufficient to learn domain-general predictors. Unlike previous work, we outline the constraints that identify the strict and complete set of causal mechanisms to achieve domain generality. We then propose regularizers to implement these constraints and empirically show the efficacy of our proposed algorithm compared to the state-of-the-art on synthetic and real-world data. To this end, we observe that the conditional independence measures are effective for model selection -- outperforming standard validation approaches. While our contributions are focused on methodology, our results also highlight existing gaps in standard evaluation using domain-average metrics -- which we show can hide worst-case performance; arguably a more meaningful measure of domain generality. 


\section{Related Work}
Domain adaptation and generalization have grown to be large sub-fields in recent years. Thus, we do not attempt an exhaustive review, and will only highlight a few papers most related to our work.
To address covariate shift, \citep{BenDavid2009ATO} gives bounds on target error based on the $\mathcal{H}$-divergence between the source and target covariate distributions, which motivates domain alignment methods like the Domain Adversarial Neural Networks \citep{ganin2016domain}. Others have followed up on this work with other notions of covariant distance for domain adaptation such as mean maximum discrepancy (MMD) \citep{long2016unsupervised} and Wasserstein distance \citep{courty2017joint}, etc. However,  \citep{pmlr-v75-kpotufe18a} show that these divergence metrics fail to capture many important properties of transferability, such as asymmetry and non-overlapping support. \citep{Zhao2019OnLI} show that even with distribution alignment of covariates, large distances between label distributions inhibit transfer; they propose a label conditional importance weighting method to address this limitation. Additionally, \cite{schrouff2022maintaining} show that many real-world problems contain more complicated `compound' shifts than covariate shifts. Furthermore, domain alignment methods are useful when one has unlabeled or partially labeled samples from the domain one would like to adapt to during training, however, the domain generalization problem setting may not include such information. The notion of invariant representations starts to address the problem of domain generalization, the topic of this work, rather than domain adaptation.

\citep{Arjovsky2019InvariantRM} propose an objective to learn a representation of the observed features which, when conditioned on, yields a domain-invariant distribution on targets, i.e., conditionally independent of domain. They argue that satisfying this invariance gives a feature representation that only uses domain-general information. However, \citep{Rosenfeld2021TheRO} shows that the IRM objective can fail to recover a predictor that does not use spurious correlations without observing a number of domains greater than the number of spurious features, which can inhibit generalization. Variants of this work \citep{Krueger2021OutofDistributionGV, wang2022provable} address this problem with higher order moment constraints to reduce the necessary number of observed domains. However, we will show that invariance on the observed domain is insufficient for domain generalization.

Additionally, one of the motivations for domain generalization is mitigating the worst domain performance. \cite{Gul2020LostDG} observe empirically that ERM is competitive and often best in worst domain accuracy across a range of real-world datasets. \cite{Rosenfeld2021AnOL} analyze the task of domain generalization as extrapolation via bounded affine transformations and find that ERM remains minimax optimal in the linear regime. However, the extent of the worst-domain shift is often unknown in practice and may not be bounded affine transformations \cite{shen2021towards}. 

In contrast, our work allows for arbitrary distribution shifts, provided that causal mechanisms remain unchanged. In addition, we show that our proposed method gives a predictor that recovers all domain-general mechanisms and is free of spurious correlations without necessitating examples (neither labeled nor unlabeled) from the target domain.

\section{Problem Setup} \label{sec:background}
\begin{figure}[t]
    \centering
    \includegraphics{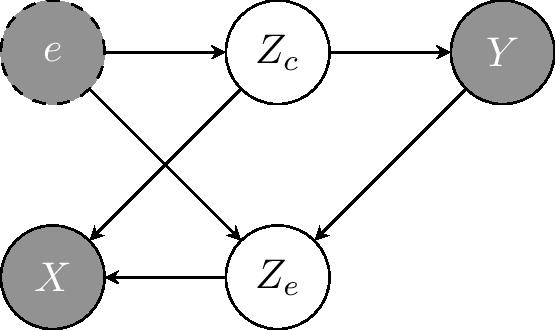}
    \caption{Graphical model depicting the structure of our data-generating process - shaded nodes indicate observed variables. $X$ represents the observed features, $Y$ represents observed targets, and $e$ represents domain influences. There is an explicit separation of domain-general (causal) $Z_c$ and domain-specific (anticausal) $Z_e$ features, which are combined to generate observed $X$.
    }
    \label{fig:graph}
\end{figure}

We consider the data generating mechanism as described by the causal graph in Figure \ref{fig:graph} and the equivalent structural equation model (or structural causal model $\mathcal{SCM}$ (equation \ref{eq:sem}; \citep{Pearl2010CausalI}). One particular setting where this graph applies is medicine, where we are often interested in predicting conditions from potential causes and symptoms of the condition. Additionally, these features may be influenced by demographic factors varying across hospitals \citep{schrouff2022maintaining}. Another setting is when measuring events in a physical process much faster than the measurement frequency; one observes both upstream (causal) and downstream (anticausal) features of the events of interest. An example is in task-fMRI where the BOLD (Blood-Oxygen-Level-Dependent) signal in task-fMRI (functional Magnetic Resonance Imaging) is much slower than the neural processes in the brain that encode the task to be predicted \citep{Glover2011OverviewOF}. Many other real-world problems fall under this causal and anticausal setting; this graph is also assumed by previous work \cite{Arjovsky2019InvariantRM}. We also assume that the observed data are drawn from a set of $E_{tr}$ training domains $\mathcal{E}_{tr} = \{e_i: i=1, 2, \ldots, E_{tr}\}$, all generated from Equation \ref{eq:sem}, thereby fixing the mechanisms by which the observed distribution is generated:

\begin{equation} \label{eq:sem}
    \mathcal{SCM}(e_i) \coloneqq
    \begin{cases}
      z_c^{(e_i)} \sim P_{Z_c}^{(e_i)}& \\
      y^{(e_i)} = f_y\left(z_c^{(e_i)}, \nu\right) & \text{where } \nu \indep z_c^{(e_i)},\\
      z_e^{(e_i)} = f_{z_e}\left(y^{(e_i)}, \eta^{(e_i)}\right) & \text{where } \eta^{(e_i)} \indep y^{(e_i)},
    \end{cases}
\end{equation}

\noindent where $P_{Z_c}$ is the causal covariate distribution, $f_y,\, f_{z_e}$ are generative mechanisms of $y$ and $z_e$, respectively, and $\nu,\, \eta$ are exogenous variables. These mechanisms are assumed to hold for any domain generated by this generative process, i.e., $\mu_{e_i}(y\,|\,z_c) = \mu(y\,|\,z_c) \text{ and } \nu_{e_i}(z_e\,|\,y) = \nu(z_e\,|\,y)\, \forall e_i\in \mathcal{E}$ for some distributions $\mu \text{ and } \nu$, where $\mathcal{E}$ is the set of all possible domains. Under the Markov assumption, we can immediately read off some properties of any distribution induced by the data-generating process shown in Figure \ref{fig:graph}: (i) $e \indep Y\, |\, Z_c$, (ii) $Z_c \indep Z_e\, |\, Y, e$, and (iii) $Y \not \indep e\, |\, X$. However, as shown in Figure \ref{fig:graph}, we only observe an unknown function of latent variables $z_c$ and $z_e$, $x = h(z_c, z_e)$ and we would like predictions for a fixed input that do not depend on domains ($e$). Consequently, a sound strategy is to map $x \rightarrow z_c$ such that the mechanism $f_y$ can be learned, as {\bf this would suffice for domain generalization.}

In contrast, though we have that the mechanism $z_e \rightarrow y$ is preserved, we have no such guarantee on the inverse $z_e \not \leftarrow y$ as it may not exist or be unique and, {\em therefore, does not satisfy domain generalization.} This is also a problem for mappings that include $z_e$:
\begin{align*}
\mu_{e_i}(y\,|\,z_e) &\ne \mu_{e_j}(y\,|\,z_e) \text{ and }\\
\mu_{e_i}(y\,|\,z_c, z_e) &\ne \mu_{e_j}(y\,|\,z_c, z_e) \text{ for } i\ne j.
\end{align*}

The latter implies that $\mu_{e_i}(y\,|\, x) \ne \mu_{e_j}(y\,|\,x) \text{ for } i\ne j$, and therefore, the original observed features will not be domain-general. Note that the generative process and its implicates are vital for our approach, since we assume that they are preserved across domains.

\begin{assumption} \label{assum:generative}
We assume that all distributions, train, and test (observed and unobserved at train), are generated by the generative process described in Equation \ref{eq:sem}.
\end{assumption}

In the following sections, we will introduce our proposed algorithm to learn a feature extractor that maps to $Z_c$ and show the generative properties that are necessary and sufficient to do so.

\section{Target-Conditioned Representation Independence Objective}
\begin{figure}[!t] 
    \centering
    \resizebox{0.75\textwidth}{!}{\includegraphics[scale=1.25]{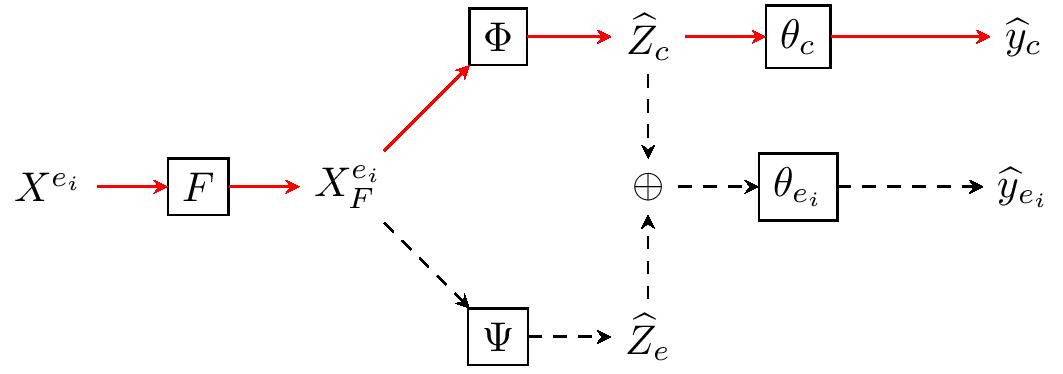}}
    \caption{We first generate a feature representation via featurizer $F$. During training, both representations, $\Phi$ and $\Psi$, generate predictions -- domain-general and domain-specific predictions, respectively. However, only the domain-invariant representations/predictions are used during test time -- indicated by the solid red arrows. $\oplus$ indicates concatenation.}
    \label{fig:architecture}
\end{figure}

We first define two distinct types of representations of observed features that we will need henceforth -- domain-invariant and domain-general.

\begin{definition} \label{def:dom-inv} A \textbf{domain-invariant} representation $\Phi(X)$, with respect to a set of observed domains $\mathcal{E}_{tr}$, is one that satisfies $\mu_{e_i}\left(y | \Phi(X)=x\right) = \mu_{e_j}\left(y | \Phi(X)=x\right) \forall e_i,\, e_j \in \mathcal{E}_{tr}$ for any fixed $x$, where $e_i,\, e_j$ are domain identifiers and $\mu$ is a probability distribution.
\end{definition}

In other words, under a domain-invariant representation, the output conditional distribution for a given input is necessarily the same across the reference (typically observed) domains. This is consistent with the existence of $\mu$ s.t. $\mu\left(y | \Phi(X)=x\right) = \mu\left(y | \text{do}(\Phi(X)=x)\right)$, where $\text{do}(\cdot)$ is the "do-operator"~\cite{Pearl2010CausalI}, denoting an intervention (arbitrary assignment of value). Conversely, for a domain-specific representation, the output conditional distribution for a given input need not be the same across domains.


However, by definition, this representation is domain-invariant up to a specific set of domains, often the set of domains that it is learned on. So, it may not be domain-invariant with respect to test domains without additional assumptions on the set of training domains, e.g., their convex hull \citep{Rosenfeld2021TheRO}. So, we need to refine this property to better specify the representations we would like to learn. This motivates the following definition, which ties domain-invariance to a specific generative process, as opposed to a set of the observed distributions, which, along with Assumption \ref{assum:generative}, connects causality to domain-generality.

\begin{definition} \label{def:dom-gen}
A representation is \textbf{domain-general} for a DAG $\mathcal{G}$ if it is domain-invariant for $\mathcal{E}$, where $\mathcal{E}$ is the set of all possible domains generated by $\mathcal{G}$.
\end{definition}

By Assumption \ref{assum:generative}, causal mechanisms from features $Z_c$ to target $Y$ are domain-general, so a natural strategy is to extract $Z_c$ from the observed features. To understand when we can recover $Z_c$ from the observed $X$, we will consider two conditional independence properties implied by the assumed causal graph: $Y \indep e\,|\, Z_c$, which we call the {\em domain invariance property}, and $Z_c \indep Z_e \,|\, Y, e$, which we capture in the following {\em target conditioned representation independence property} (TCRI; definition \ref{def:tcri}).

\begin{definition}[Target Conditioned Representation Independence] \label{def:tcri}
Two functions, $\Phi,\, \Psi$, are said to satisfy TCRI with respect to random variables $X,\, Y,\, e_i$ if $I(\Phi(X), \Psi(X); Y) = I(Z_c, Z_e; Y)$
(total-chain-information-criterion) and $\Phi(X) \indep \Psi(X) \,|\, Y \,\forall e_i$.
\end{definition}

We will show in Section \ref{sec:theory} that these properties (domain invariance and TCRI) together identify $Z_c$ from $X, Y$ to give a domain-general representation. Based on our results, we design an algorithm for learning a feature mapping that recovers $Z_c$, i.e., the domain-general representation (mechanisms). Figure \ref{fig:architecture} illustrates the learning framework. 

In practice, we propose a TCRI objective containing four terms, each related to the properties desired of the learned representations, as follows,
\begin{equation} \label{eq:ours_L}
    \mathcal{L}_{TCRI} = \frac{1}{E_{tr}}\sum_{e_i\in\mathcal{E}_{tr}}\left[\alpha \mathcal{L}_{\Phi} + (1 - \alpha) \mathcal{L}_{\Phi \oplus \Psi} + \lambda \mathcal{L}_{IRMv1'} + \beta \mathcal{L}_{CI} \right],
\end{equation}
where $\alpha \in [0, 1]$, $\beta > 0$, and $\lambda > 0$ are hyperparameters -- Figure \ref{fig:architecture} shows the full framework.

In detail, we let $\mathcal{L}_\Phi = \mathcal{R}(\theta_c \cdot \Phi)$ represent the domain-general predictor accuracy and let $\mathcal{L}_{IRMv1'}$ be a penalty from on $\Phi$ and the linear predictor $\theta_c$ that enforces that $\Phi$ has the same optimal predictor across training domains, capturing the domain invariance property, where $\Phi: \mathcal{X} \mapsto \mathcal{H}_\Phi$, $\theta_c: \mathcal{H}_\Phi \mapsto \mathcal{Y}$ \citep{Arjovsky2019InvariantRM}:
\begin{equation*}
    \mathcal{L}_{IRMv1'} = \|\nabla_{\theta_c} \mathcal{R}^{e_i}(\theta_c \circ \Phi)\|_2,
\end{equation*}
where $\mathcal{R}^{e_i}$ denotes the empirical risk achieved on domain $e_i$. The $\mathcal{L}_\Phi$ and $\mathcal{L}_{IRMv1'}$ together implement a domain-invariance property, however, we know that this is not sufficient for domain generalization \citep{Rosenfeld2021TheRO}. We will show later that the addition of the TCRI property suffices for domain generalization (Theorem \ref{thm:irm_tcri}).

To implement the TCRI property (definition \ref{def:tcri}), we also learn a domain-specific representation $\Psi$, which is constrained to be (i) conditionally independent of the domain-general representation given the target and domain and (ii) yield a predictor as good as one from $X$ when combined with $\Phi$. We first address (ii); given a domain-specific representation, $\Psi: \mathcal{X} \mapsto \mathcal{H}_\Psi$, we define a set of domain-specific predictors $\{\theta_{e_i}: \mathcal{H}_\Phi \times \mathcal{H}_\Psi \mapsto \mathcal{Y}: i=1,\ldots,E_{tr}\}$. We then add a term in the objective that minimizes the loss of these domain-specific predictors:
\begin{equation*}
    \mathcal{L}_{\Phi \oplus \Psi} = \mathcal{R}^{e_i}\left(\theta_{e_i} \circ (\Phi \oplus \Psi)\right).
\end{equation*}

\noindent This term aims to enforce the total-chain-information-criterion of TCRI, by allowing $\Phi$ and $\Psi$ to minimize a domain-specific loss together, where spurious information in a domain can be used to improve within-domain performance. Since we have that both the domain-general and domain-specific have unique information about the target, the optimal model will use both types of information. We allow the $\theta_{e_i}$ for each domain to be different since, by definition of the problem, we expect these mechanisms to vary across domains.

To address (I), we define $\mathcal{L}_{CI}$ to be the conditional independence part of the TCRI property and use the V-statistic-based Hilbert-Schmidt Independence Criterion (HSIC) estimate (\citep{Gretton2007AKS}). For the two representations $\Phi(X),\, \Psi(X)$, define conditional independence ($\Phi(X) \indep \Psi(X) \,|\, K$) as
\begin{align*}
    \mathcal{L}_{CI} = \frac{1}{C}\sum_{k=1}^C \widehat{HSIC}(\Phi(X)_k,\Psi(X)_k) = \frac{1}{C}\sum_{k=1}^C \frac{1}{n_k^2}\text{trace}(\mathbf{K}_{\Phi}\mathbf{H}_{n_k}\mathbf{K}_{\Psi}\mathbf{H}_{n_k}),
\end{align*}

\noindent where $k$, indicates which class the examples in the estimate correspond to, $C$ is the number of classes, $\mathbf{K}_{\Phi}\in \mathbb{R}^{n_k\times n_k},\, \mathbf{K}_{\Psi}\in \mathbb{R}^{n_k\times n_k}$ are Gram matrices, $\mathbf{K}_{\Phi}^{i,j} = \kappa(\Phi(X)_i,\Phi(X)_j)$, $\mathbf{K}_{\Psi}^{i,j} = \omega(\Psi(X)_i, \Psi(X)_j)$ with kernels $\kappa, \omega$ are radial basis functions, $\mathbf{H}_{n_k} = \mathbf{I}_{n_k} - \frac{1}{n_k^2}\mathbf{1}\mathbf{1}^\top$ is a centering matrix, $\mathbf{I}_{n_k}$ is the ${n_k} \times {n_k}$ dimensional identity matrix, $\mathbf{1}_{n_k}$ is the ${n_k}$-dimensional vector whose elements are all 1, and $^\top$ denotes the transpose. We condition on the label by taking only examples of each label and computing the empirical HSIC; then, we take the average. We note that any criterion for conditional independence can be used as $\mathcal{L}_{CI}$, e.g., partial covariance.

Altogether, we have the following objective:
\begin{align} \label{eq:tcri_min}
    \min_{\Phi,\Psi, \theta_c, \theta_1, \theta_2, \ldots,\theta_{E_{tr}}} \frac{1}{E_{tr}}\sum_{e_i\in\mathcal{E}_{tr}} \left[\alpha\mathcal{L}_{\Phi} + (1 - \alpha) \mathcal{L}_{\Phi \oplus \Psi} + \mathcal{L}_{IRMv1'} + \beta \mathcal{L}_{CI}\right].
\end{align}

\noindent We compute the complete objective for each domain separately to condition on the domains $e$, and after minimizing this objective, only the invariant representation and its predictor, $\theta_c \circ \Phi$, are used.

\section{Conditions for a Domain-General Representations} \label{sec:theory}
Now we provide some analysis to justify our method. Consider a feature extractor $\Phi: \mathcal{X} \mapsto \mathcal{Z}$, where $\mathcal{X}, \mathcal{Z}$ are input and latent features spaces, respectively.
We first show that a $\Phi$ that captures a strict subset of the causal features can satisfy the domain invariance property $Y \indep e \,|\, \Phi(X)$ while not necessarily being domain-general -- Lemmas \ref{lem:subset} and \ref{lem:irm} (proofs in Appendix \ref{sec:proof}).

\begin{lemma}[Insufficiency of Causal Subsets for domain generalization] \label{lem:subset}
 Conditioning on a subset of causal variables (invariant mechanisms) does not imply domain generalization (definition \ref{def:dom-gen}). \[\mathcal{Z} \subset \mathcal{Z}_c \notimplies \mu_{e_i}(y \,|\, Z=z) = \mu_{e_j}(y \,|\, Z =z) \forall e_i \ne e_j,\, z \in \mathcal{Z}\]
where $\mathcal{Z}_c$ is the causal feature space. (Proof in Appendix \ref{sec:proof})

\end{lemma}

\begin{lemma} \label{lem:irm}
A representation $\Phi$ that maps to a strictly causal subset can be domain-invariant.  (Proof in Appendix \ref{sec:proof})
\end{lemma}

Specifically, we show that causal subsets can satisfy the domain-invariance property ($\Phi(X) \indep Y \,|\, e$), but an incomplete causal representation may not be domain-general if the shifts at test-time are on causal features that are not captured in the incomplete causal representation. To address this, we show that the domain-invariance property and Target Conditioned Representation Independence (TCRI) property together are sufficient for recovering the complete set of causal (invariant) mechanisms.

\begin{lemma} \label{lem:aggregation}
(Sufficiency of TCRI for Causal Aggregation). Recall, $X,\, Z_c,\, Z_e,\, Y$ from Figure \ref{fig:graph}. Let $Z_c, Z_e$ be direct causes and direct effects of $Y$, respectively, and recall that $X$ is a function of $Z_c$ and $Z_e$. If the two representations induced by feature extractors $\Phi, \Psi$ satisfy TCRI, then wlog $I(\Phi(X);  Y) \ge I(Z_c; Y)$.
\end{lemma}

Lemma \ref{lem:aggregation} (proof in Appendix \ref{sec:theory}) shows that the TCRI property addresses the limitation identified in Lemma \ref{lem:subset}, that is, the TCRI property implies that all causal information about the target is aggregated into one representation $\Phi$. Now, to identify $Z_c$, we only need to show that $\Phi$ is strictly causal.

\begin{theorem}[Sufficiency of TCRI + domain-invariance for identifying $Z_c$] \label{thm:irm_tcri}
Recall that $Z_c, Z_e$ are the true latent features in the assumed generative model and $X$ are the observed features -- Figure \ref{fig:graph}. If $\Phi,\, \Psi$ satisfy TCRI and domain-invariance, then $\Phi$ recovers $Z_c$ and is therefore domain-general.

\begin{proof}
By Lemma \ref{lem:aggregation}, we have that when $\Phi, \Psi$ satisfy TCRI, $\Phi(X)$ contains all of the causal (necessarily domain-general) information in $X$. However, $\Phi(X)$ may also contain non-causal (domain-specific) information, spoiling the domain-generality of $\Phi$. It remains to show that $\Phi(X)$ is strictly causal when we add the domain-invariance property.

\noindent If $\Phi$ satisfies the domain-invariance property, then $Y \indep e \,|\, \Phi(X)$. Clearly, this cannot be the case if $\Phi(X)$ contains features of $Z_e$, since $e$ and $Y$ are colliders on $Z_e$ and therefore conditioning on $Z_e$ opens a path between $e$ and $Y$, making them dependent. Thus $\Phi(X)$ can only contain causal features.

\noindent Therefore, a representation that satisfies TCRI and the domain-invariance property is wholly and strictly causal and thus domain-general. The latter follows from the $Z_c$ having invariant mechanisms.
\end{proof}
\end{theorem}

Theorem \ref{thm:irm_tcri} suggests that by learning two representations that together capture the mutual information between the observed $X,\, Y$, where one satisfies the domain-invariance property and both satisfy TCRI, one can recover the strictly and complete causal feature extractor and domain-general predictor.

\begin{remark}
    One limitation of TCRI is a failure mode when the strictly anticausal representation gives a domain invariant predictor. In this case, either representation may be $\Phi$. However, one of the benefits of having a domain-specific predictor for each observed domain is that one can check if they are interchangeable. Specifically, in this scenario, one will observe that the domain-specific classifiers give similar results when applied to a domain they were not trained on since they are based on invariant causal mechanisms. This, however, gives a test, not a fix for this setting -- we leave a fix for future work.
\end{remark}

\section{Experiments}
To evaluate our method in a setting that exactly matches our assumptions and we know the ground truth mechanisms, we use Equation \ref{eq:simsem} to generate our simulated data, with domain parameters $\sigma^{e_i},\, \sigma_\eta^{e_i}$ -- code provided in the supplemental materials.

\begin{minipage}{0.43\textwidth}
\begin{flalign} \label{eq:simsem}
    \mathcal{SCM}(e_i) \coloneqq
    \begin{cases}
      z_c^{(e_i)} \sim \text{Exp}\left(\sigma^{e_i}\right)& \\
      y^{(e_i)} = z_c^{(e_i)} + \text{Exp}\left(0.25\right),\\
      z_e^{(e_i)} = y^{(e_i)} + \text{Exp}\left(\sigma_\eta^{e_i}\right).
    \end{cases}
\end{flalign}
\end{minipage}
\hfill
\begin{minipage}{0.475\textwidth}
    \renewcommand{\arraystretch}{1.1}
    \centering
    \input{tables/simulated}
    \captionof{table}{Continuous Simulated Results -- Feature Extractor with a dummy predictor $\theta_c = 1.$, i.e., $\hat{y} = x \times \Phi \times w$, where $x\in \mathbb{R}^{N \times 2},\, \Phi \in \mathbb{R}^{2 \times 1},\, w \in \mathbb{R}$. Oracle indicates the coefficients achieved by regressing $y$ on $z_c$ directly.}
\label{tab:sim}
\end{minipage}
~\\~\\
We observe 2 domains with parameters $\sigma^{e_i=0} = \sigma_\eta^{e_i=0} = 0.1$, $\sigma^{e_i=1} = \sigma_\eta^{e_i=1} = 1$, 1000 samples, and linear feature extractors and predictors. Minimizing the TCRI objective (Equation \ref{eq:tcri_min}) recovers a linear feature representation that maps back to $z_c$ (Table \ref{tab:sim}). Note that for ERM, $\lambda = \beta = 0,\, \alpha = 1$, IRM, $\lambda = 0.1,\, \beta = 0 ,\, \alpha = 1$, and TCRI, $\lambda = 0.1, \, \beta = 10\text{ and } \alpha = 0.75$; additional details in Appendix \ref{app:sim}.
\subsection{Real-World Datasets}

\textbf{Algorithms:} We compare our method to the following baselines: Empirical Risk Minimization (\textbf{ERM}, \citep{NIPS1991_ff4d5fbb}), Invariant Risk Minimization (\textbf{IRM} \citep{Arjovsky2019InvariantRM}), Variance Risk Extrapolation (\textbf{V-REx}, \cite{Krueger2021OutofDistributionGV}), Meta-Learning for Domain Generalization (\textbf{MLDG}, \cite{li2018learning}), Group Distributionally Robust Optimization (\textbf{GroupDRO}, \citep{sagawa2019distributionally}), and Adaptive Risk Minimization (\textbf{ARM} \citep{zhang2021adaptive}).

We evaluate our proposed method on real-world datasets. Given observed domains $\mathcal{E}_{tr} = \{e_i: i=1,2,\ldots,E_{tr}\}$, we train on $\mathcal{E}_{tr} \, \backslash \, e_i$ and evaluate the model on the unseen domain $e_i$, for each $e_i$.

\textbf{Model Selection:} Typically, ML practitioners use a within-domain hold-out validation set for model selection. However, this strategy is biased towards the empirical risk minimizer, i.e., the one with the lowest error on the validation set from the training domains; however, we know that the model that achieves the highest validation accuracy may not be domain-general. This same is true if we use an out-of-domain validation set that is not from the target domain. Alternatively, we propose to leverage the generative assumptions for model selection. We consider other properties of our desired model for model selection, specifically, a low $\mathcal{L}_{CI}$. To do this, we follow the practice of a hold-out within-domain validation set; however, we compute $\mathcal{L}_{CI}$ for the validation set and choose the example with the lowest CI score instead of the highest validation accuracy. We compare this strategy with validation accuracy in our results. Additional details can be found in Appendix \ref{app:model_selection}

\textbf{ColoredMNIST:} We evaluate our method on the ColoredMNIST dataset \cite{Arjovsky2019InvariantRM} which is composed of $7000$ ($2\times 28 \times 28$, $1$) images of a hand-written digit and binary-label pairs. There are three domains with different correlations between image color and label, i.e., the image color is spuriously related to the label by assigning a color to each of the two classes (0: digits 0-4, 1: digits 5-9). The color is then flipped with probabilities $\{0.1, 0.2, 0.9\}$ to create three domains, making the color-label relationship domain-specific because it changes across domains. There is also label flip noise of $0.25$, so we expect that the best we can do is 75\% accuracy. As in Figure \ref{fig:graph}, $Z_c$ corresponds to the original image, $Z_e$ the color, $e$ the label-color correlation, $Y$ the image label, and $X$ the observed colored image. Code (a variant of \href{https://github.com/facebookresearch/DomainBed}{https://github.com/facebookresearch/DomainBed}) can be found at \href{https://github.com/olawalesalaudeen/DomainBed/tree/master/domainbed}{https://github.com/olawalesalaudeen/DomainBed/tree/master/domainbed}.

\textbf{Worst-Case--PACS} -- \emph{Variables.} $X$: images, $Y$: non-urban (elephant, giraffe, horse) vs. urban (dog, guitar, house, person). \emph{Domains.} \{\{cartoon,  art painting\}, \{art painting,  cartoon\}, \{photo\}\} \citep{li2017deeper}. The photo domain is the same as in the original dataset. In the \{cartoon, art painting\} domain, urban examples are selected from the original cartoon domain, while non-urban examples are selected from the original art painting domain. In the \{art painting, cartoon\} domain, urban examples are selected from the original art painting domain, while non-urban examples are selected from the original cartoon domain. Here, the model may use spurious correlations (domain-related information) to predict the labels; however, since these relationships are flipped between domains \{\{cartoon,  art painting\} and \{art painting,  cartoon\}, these predictions will be wrong when generalized to other domains.

\textbf{Worst-Case--VLCS} -- \emph{Variables.} $X$: images, $Y$: animate (bird, dog, person) vs. inanimate (car, chair). \emph{Domains.} \{\{SUN09, LabelMe\}, \{LabelMe, SUN09\}, \{VOC2007\}\} \citep{Fang2013UnbiasedML}. The VOC2007 domain is the same as in the original dataset. In the \{SUN09, LabelMe\} domain, animate samples are selected from the SUN09 domain, while inanimate samples are selected from the LabelMe domain. Conversely, in the \{LabelMe, SUN09\} domain, animate samples are selected from the LabelMe domain, while inanimate samples are selected from the SUN09 domains.






\subsection{Results and Discussion}
\begin{table}[!t]
\renewcommand{\arraystretch}{1.1}
\centering
\caption{Colored MNIST. `ci' indicates model selection via validation conditional independence. `cov' indicates a conditional cross-covariance penalty, and `HSIC' indicates a Hilbert-Schmidt Independence Criterion penalty. Columns \{+90\%, +80\%, -90\%\} indicate domains -- $\{0.1, 0.2, 0.9\}$ digit label and color correlation, respectively. We also show the mean, standard deviation, and min of the average domain accuracies over three trials.}
\input{tables/cmnist}
\label{tab:cmnist}
\end{table}
\textbf{Worst-domain Accuracy:} A critical implication of a domain-general is stability -- robustness in worst-domain performance, up to domain difficulty. While average accuracy across domains provides some insight into an algorithm's ability to generalize to new domains, it is susceptible to being dominated by the performance of subsets of domains. For example, ARM outperforms the baselines in average accuracy, but this improvement is driven primarily by the first domain (+90\%), while the worst-domain accuracy stays the same. In real-world challenges such as algorithmic fairness, comparable worst-domain accuracy is necessary \citep{hardt2016equality}. 

We observe that TCRI is competitive in average accuracy with the baseline methods. However, it is significantly more stable when using conditional independence for model selection, i.e., the worst-domain accuracy is highest, and variance across domains is lowest for TCRI -- both by a large margin. We note that, for ColoredMNIST (Table \ref{tab:cmnist}), domain -90\%, which has a color-label relationship flip probability of 0.9, has a majority color-label pairing opposite of domains +80\% and +90\%, with flip probabilities of 0.1 and 0.2, respectively. Hence, we observe that the baseline algorithms generalize poorly to domain -90\%, relative to the other two domains. This indicates that, unlike in TCRI, the baselines use spurious information (color) for prediction. While TCRI does not obtain the expected best accuracy of 75\%, it is evident that the information used for prediction is general across the three domains, given the low variance in cross-domain accuracy. We observe the same trends in the Worst-Case--PACS (Table \ref{tab:pacs_v3}) and Worst-Case--VLCS (Table \ref{tab:vlcs_v3}) datasets -- TCRI achieves the best worst-case accuracy while remaining competitive in average accuracy (highest average-case also for Worst-Case--PACS).

Additionally, TCRI achieves the lowest standard deviation across domain accuracies for all datasets, showing further evidence of the cross-domain stability one would expect from a domain-general model. The baselines also include V-REx, which implements explicit regularizers on risk variance across observed domains, but TCRI achieves a much lower variance for domain accuracies.

Furthermore, we provide additional empirical results on ColoredMNIST, which illustrates the effect of similarities between source and target domain distributions on accurate evaluation domain generality in Appendix \ref{sec:benchmark}. We also report Oracle accuracies to highlight further the soundness of the TCRI regularization and model selection approach, showing that it does not reject good models.

\begin{table}[t]
    \centering
    \caption{Worst-Case--PACS. Domain accuracies and all-domain statistics. `ci' indicates model selection via the lowest conditional independence score (as in the objective term) on a held-out source domains set. Results highlight that the proposed TCRI achieves the best average-case and worst-case domain performance.}
    \label{tab:pacs_v3}
    \renewcommand{\arraystretch}{1.1}
    \input{tables/pacs_v3}
\end{table}

\begin{table}[t]
    \centering
    \caption{Worst-Case--VLCS. Domain accuracies and all-domain statistics. `ci' indicates model selection via the lowest conditional independence score (as in the objective term) on a held-out source domains set. Results highlight that the proposed TCRI achieves the best worst-case domain performance.}
    \label{tab:vlcs_v3}
    \renewcommand{\arraystretch}{1.1}
    \input{tables/vlcs_v3}
\end{table}

\textbf{Limitations:} The strength of TCRI is also its limitation; TCRI is very conservative to be robust to worst-domain shifts. While many critical real-world problems require robustness to worst-domain shifts, this is not always the case, and in this setting, TCRI sacrifices performance gains from non-domain-general information that may be domain-invariant for the expected domains. The practitioner should apply this method when appropriate, i.e., when domain generality is critical and the target domains may differ sufficiently from the source domain. It is, however, essential to note that in many settings where one is happy with domain-invariance as opposed to domain-generality, ERM may be sufficient \citep{Gul2020LostDG, Rosenfeld2021AnOL}.

\section{Conclusion and Future Work}
We address the limitations of state-of-the-art algorithms' inability to learn domain-general predictors by developing an objective that enforces DAG properties sufficient to disentangle causal (domain-general) and anticausal (domain-specific) mechanisms. We show that our method is competitive with other state-of-the-art domain-generalization algorithms on real-world datasets in terms of average across domains. Moreover, TCRI outperforms all baseline algorithms in worst-domain accuracy, indicating the expected stability across domains that one would expect from domain-general predictors. We also find that using conditional independence metrics for model selection outperforms the typical validation accuracy strategy. Future work includes further investigating other model selection strategies that preserve the desired domain-generality properties and curating more benchmark real-world datasets that exhibit worst-case behavior.\\

\noindent \textbf{Acknowledgements:} This research was supported by the National Science Foundation under grant 1735252.

\bibliographystyle{plainnat}
\bibliography{main}

\newpage
\appendix
\input{99_appendix}

\end{document}

%% file: tables/simulated.tex
\begin{tabular}{c|c|c}
     \textbf{Model} & \textbf{$\Phi_{0,0}$} & \textbf{$\Phi_{1,0}$} \\
     \hline
     ERM            & 0.84 & 0.18 \\
     IRM            & 0.83 & 0.18 \\
     TCRI (HSIC)    & 1.11 & 0.01\\
     \hline
     Oracle         & 1.13   & 0.0 \\
\end{tabular}

%% file: tables/cmnist.tex
\begin{tabular}{l|c|c|c?c|c|c}
&
\multicolumn{3}{c?}{Domains} &
\multicolumn{3}{c}{Domain Accuracy Statistics}\\
\hline
\textbf{Algorithm} & \textbf{+90\%} & \textbf{+80\%} & \textbf{-90\%} & \textbf{mean} & \textbf{std} & \textbf{min}\\
\hline
ERM                         & 71.8 $\pm$ 0.1       & 72.8 $\pm$ 0.2       & 10.0 $\pm$ 0.1       & 51.5                 & 36.0       & 10.0 \\  
IRM                         & 72.4 $\pm$ 0.5       & 72.8 $\pm$ 0.2       & 10.0 $\pm$ 0.3       & 51.7                 & 36.1       & 10.0 \\  
VREx                        & 72.3 $\pm$ 0.4       & 73.2 $\pm$ 0.5       & 10.0 $\pm$ 0.1       & 51.8                 & 36.2       & 10.0 \\
GroupDRO                    & 72.4 $\pm$ 0.2       & 73.1 $\pm$ 0.2       & 10.0 $\pm$ 0.2       & 51.8                 & 36.2       & 10.0 \\  
MLDG                        & 71.7 $\pm$ 0.0       & 73.0 $\pm$ 0.1       & 10.2 $\pm$ 0.0       & 51.6                 & 35.9       & 10.2 \\  
ARM                         & 81.9 $\pm$ 0.6       & 74.5 $\pm$ 1.2       & 10.2 $\pm$ 0.0       & 55.6                 & 39.4       & 10.2 \\  
 
\hline 
TCRI (cov) -- ci            & 54.7 $\pm$ 1.1       & 56.4 $\pm$ 2.4       & 50.6 $\pm$ 0.1       & 53.9                 & \textbf{3.0}       & 50.6 \\
TCRI (HSIC) -- ci           & 54.7 $\pm$ 1.6       & 60.1 $\pm$ 4.1       & 53.0 $\pm$ 2.1       & \textbf{55.9}                 & 3.6        & \textbf{53.0}
\end{tabular}

%% file: tables/pacs_v3.tex
\begin{tabular}{l|c|c|c?c|c|c}
&
\multicolumn{3}{c?}{Domains} &
\multicolumn{3}{c}{Domain Accuracy Statistics}\\
\hline
\textbf{Algorithm} & \textbf{C x A} & \textbf{A x C} & \textbf{P} & \textbf{mean} & \textbf{std} & \textbf{min}\\
\hline
ERM                  & 31.2 $\pm$ 1.3       & 42.8 $\pm$ 0.7       & 97.6 $\pm$ 0.2       & 57.2                 & 29.0                 & 31.2                 \\
IRM                  & 30.3 $\pm$ 0.3       & 39.0 $\pm$ 1.3       & 94.9 $\pm$ 1.4       & 54.7                 & 28.6                 & 30.3                 \\
GroupDRO             & 37.7 $\pm$ 0.7       & 42.1 $\pm$ 1.6       & 95.7 $\pm$ 0.5       & 58.5                 & 26.4                 & 37.7                 \\
MLDG                 & 34.9 $\pm$ 2.4       & 41.7 $\pm$ 2.2       & 96.8 $\pm$ 0.3       & 57.8                 & 27.7                 & 34.9                 \\
ARM                  & 34.1 $\pm$ 0.8       & 43.8 $\pm$ 1.1       & 96.5 $\pm$ 0.5       & 58.1                 & 27.4                 & 34.1                 \\
VREx                 & 37.5 $\pm$ 1.1       & 43.0 $\pm$ 0.5       & 95.7 $\pm$ 1.5       & 58.8                 & 26.2                 & 37.5                 \\
 
\hline 
TCRI (cov) -- ci            & 62.8 $\pm$ 0.1       & 62.3 $\pm$ 0.2       & 65.0 $\pm$ 0.4       & \textbf{63.4}                 & \textbf{1.2}                  & \textbf{62.3}                 \\
TCRI (HSIC) -- ci           & 35.1 $\pm$ 2.0       & 52.5 $\pm$ 4.5       & 68.5 $\pm$ 11.5      & 52.0                 & 13.7                 & 35.1                 \\
\end{tabular}

%% file: tables/vlcs_v3.tex
\begin{tabular}{l|c|c|c?c|c|c}
&
\multicolumn{3}{c?}{Domains} &
\multicolumn{3}{c}{Domain Accuracy Statistics}\\
\hline
\textbf{Algorithm} & \textbf{S x L} & \textbf{L x S} & \textbf{V} & \textbf{mean} & \textbf{std} & \textbf{min}\\
\hline
ERM                  & 37.9 $\pm$ 2.8       & 47.1 $\pm$ 0.8       & 82.7 $\pm$ 1.7       & 55.9                 & 19.3                 & 37.9                 \\
IRM                  & 41.3 $\pm$ 1.6       & 46.2 $\pm$ 1.2       & 68.6 $\pm$ 2.6       & 52.1                 & 11.9                 & 41.3                 \\
GroupDRO             & 37.6 $\pm$ 1.1       & 50.9 $\pm$ 1.6       & 84.2 $\pm$ 0.6       & \textbf{57.6}                 & 19.6                 & 37.6                 \\
MLDG                 & 35.9 $\pm$ 1.3       & 48.7 $\pm$ 1.3       & 81.7 $\pm$ 1.9       & 55.4                 & 19.3                 & 35.9                 \\
ARM                  & 33.4 $\pm$ 1.1       & 44.7 $\pm$ 0.8       & 83.6 $\pm$ 2.5       & 53.9                 & 21.5                 & 33.4                 \\
VREx                 & 38.1 $\pm$ 2.5       & 43.5 $\pm$ 0.8       & 80.0 $\pm$ 0.6       & 53.9                 & 18.6                 & 38.1                 \\
 
\hline 
TCRI (cov) -- ci            & 49.6 $\pm$ 0.4       & 50.4 $\pm$ 0.1       & 55.7 $\pm$ 9.0       & 51.9                 & \textbf{2.7}                  & \textbf{49.6}                 \\
TCRI (HSIC) -- ci           & 49.5 $\pm$ 0.3       & 43.9 $\pm$ 3.9       & 60.6 $\pm$ 2.7       & 51.3                 & 6.9                  & 43.9                 \\
\end{tabular}

%% file: 99_appendix.tex
\section{Simulated Data} \label{app:sim}
We observe 2 domain with parameters $\sigma^{e_i=0} = \sigma_\eta^{e_i=0} = 0.1$, $\sigma^{e_i=1} = \sigma_\eta^{e_i=1} = 1$, each with 1000 samples. We let , and use linear feature extractors and predictors. Minimizing the TCRI objective (Equation \ref{eq:tcri_min}) recovers a linear feature representation that maps back to $z_c$ (Table \ref{tab:sim}). Note that for ERM, $\lambda = \beta = 0,\, \alpha = 1$, IRM, $\lambda = 0.1,\, \beta = 0 ,\, \alpha = 1$, and TCRI, $\lambda = 0.1, \, \beta = 10\text{ and } \alpha = 0.75$; additional details can be found in Appendix \ref{app:sim}.

In addition to letting $\theta_c = 1.$ be a dummy variable, we also solve the OLS (Ordinary Least Squares) problem to compute the $\mathcal{L}_{\Phi \oplus \Psi}$ term in the loss. Each backward pass takes in all examples from a domain as a batch.

\section{Models and Hyperparameter Selection}
We do a random search over hyperparameters for our method -- four randomly selected hyperparameter sets in total. Additionally, we run three trials for each set to generate standard errors. Additional sampling details can be found in \href{https://anonymous.4open.science/r/DomainBed-8D3F/domainbed/hparams_registry.py}{https://anonymous.4open.science/r/DomainBed-8D3F/domainbed/hparams\_registry.py}. We use the default values for the baseline algorithms since the results closely match those reported by \citet{Gul2020LostDG}. The hyperparameters used for each trial are also provided in the supplemental material. 

\subsection{ColoredMNIST}
We use MNIST-ConvNet \cite{Gul2020LostDG} backbones for the MNIST datasets and parameterize our experiments with the DomainBed hyperparameters with three trials to select the best model \cite{Gul2020LostDG}. The MNIST-ConvNet backbone corresponds to the generic featurizer $F$ in Figure \ref{fig:architecture}, and both $\Phi$ and $\Psi$ are linear layers of size $128 \times 128$ that are appended to the backbone. The predictors $\theta_c,\, \theta_1,\, \ldots,\,\theta_{E_{tr}}$ are also parameterized to be linear and appended to the $\Phi$ and $\Psi$ layers, respectively.

\subsection{Worst-Case--Datasets}
We use a ResNet \cite{he2016deep} backbone to be the generic featurizer $F$, and both $\Phi$ and $\Psi$ are linear layers of size $2048 \times 2048$ that are appended to the backbone. The linear predictors, a domain-general one and one for each training domain, are appended to the $\Phi$ and $\Psi$ layers, respectively. We select hyperparameters in the same as the previous section.

\section{Model Selection} \label{app:model_selection}
Across the hyperparameter sweep, we select the model with the lowest average conditional independence score between the two TCRI representations (Definition \ref{def:tcri}) to evaluate on our test set. This is in lieu of selecting the model with the highest validation accuracy on the training domains -- we know that this is the model selection heuristic for the baselines.

Additionally, we show results based on oracle selection, that is, selection based on held-out target domain data. We observe that TCRI still outperforms the baseline methods in worst-case accuracy and has average accuracies that are competitive with the baselines. This suggests that the regularizers are not so harsh that the TCRI models cannot learn good predictors in practice. The results, however, do suggest that there is room for improvement in model selection. We leave this for future work.

\begin{table}[h]
        \centering
        \caption{Colored MNIST. Columns \{+90\%, +80\%, -90\%\} indicate domains -- $\{0.1, 0.2, 0.3, 0.9\}$ digit label and color correlation, respectively. We report the mean, standard deviation, and minimum of the average domain accuracies over 3 trials each. All models use the oracle selection method -- held out target data.}
        \input{tables/cmnist_oracle}
        \label{tab:cmnist_oracle}
    \end{table}

\section{On Benchmark Datasets for Evaluating Domain Generalization -- Worst-Case} \label{sec:benchmark}
We show some results below that illustrate the challenge of accurately evaluating the efficacy of an algorithm in domain generalization. We first note that we expect ERM (naive) to perform poorly in domain generalization tasks, certainly so when we observe worst-case shifts at test time. However, like other works \citep{Gul2020LostDG}, we observe that ERM performs as well as other baselines during transfer on various benchmark datasets. Previous theoretical results \citep{Rosenfeld2021AnOL} suggest that this observation is indicative of properties of the benchmark domains that may be sufficient for ERM to be minimax optimal - specifically that the distribution (and equivalently the loss) of the target domain can be written as a convex combination of the those in the source domains.

To further investigate this, we develop additional experiments motivated by the ColoredMNIST \citep{Arjovsky2019InvariantRM} which seems to not fall into the scenario in \citep{Rosenfeld2021AnOL}. We note that in the +90\%, +80\%, and -90\% domains of ColoredMNIST, the -90\% domain has the opposite relationship between the spurious correlation and the label, so the use of spurious correlation generalizes catastrophically in the -90\% domain. In the setting, the baseline algorithms we present achieve poor accuracy in the -90\% domain while maintaining high accuracy in the +90\% and +80\% domains. Consequently, we investigate two settings, \emph{setting a}: +90\%, +80\%, +70\%, -90\% domains and \emph{setting b}: +90\%, +80\%, -80\%, -90\% domains. In setting a, we add another domain with the majority direction in the relationship between spurious correlation and labels. In setting b, we add another domain with the minority direction.

We use Oracle model selection (held-out target data) to remove the effect of model selection for all methods in the results. We find that in setting a, where we add a domain (+70\%) that has spurious correlations that do not generalize the -90\% domain, we observe worst-case accuracy across baselines is still very different from the median-case \ref{tab:cmnist3_oracle}).

\begin{table}[h]
    \centering
    \renewcommand{\arraystretch}{1.1}
    \caption{Colored MNIST \emph{setting a}. Columns \{+90\%, +80\%, +70\%, -90\%\} indicate domains -- $\{0.1, 0.2, 0.3, 0.9\}$ digit label and color correlation, respectively. We report domain accuracies over 3 trials each. We use the oracle selection method -- held out target data.}
    \input{tables/cmnist3_oracle}
    \label{tab:cmnist3_oracle}
\end{table}

However, in setting b, where we add a domain (-80\%) that has spurious correlations that generalize to the -90\% domain, we observe that the worst-case accuracy is much closer than the median-case -- single digit standard deviation across domains \ref{tab:cmnist8_oracle}.
\begin{table}[h]
    \centering
    \renewcommand{\arraystretch}{1.1}
    \caption{Colored MNIST \emph{setting b}. Columns \{+90\%, +80\%, -80\%, -90\%\} indicate domains -- $\{0.1, 0.2, 0.8, 0.9\}$ digit label and color correlation, respectively. We report the average domain accuracies over 3 trials each. We use the oracle selection method -- held out target data.}
    \input{tables/cmnist8_oracle}
    \label{tab:cmnist8_oracle}
\end{table}

\section{Real-World Dataset} \label{sec:realworlddata}
We evaluate our methods on real-world datasets with the worst-case shifts we aim to be robust to.

\subsection{Worst-Case--PACS}
\textbf{Data description:} $X$: images, $Y$: non-urban (elephant, giraffe, horse) vs. urban (dog, guitar, house, person).

Domains: \{\{cartoon,  art painting\}, \{art painting,  cartoon\}, \{photo\}\} \citep{li2017deeper}. The photo domain is the same as in the original dataset. In the \{cartoon, art painting\} domain, urban examples are selected from the original cartoon domain, while non-urban examples are selected from the original art painting domain. In the \{art painting, cartoon\} domain, urban examples are selected from the original art painting domain, while non-urban examples are selected from the original cartoon domain. Here, the model may use spurious correlations (domain-related information) to predict the labels; however, since these relationships are flipped between domains \{\{cartoon,  art painting\} and \{art painting,  cartoon\}, these predictions will be wrong when generalized to other domains.

\begin{table}[h]
    \centering
    \caption{Worst-Case--PACS. Domain accuracies and all-domain statistics. ci indicates model selection via the lowest conditional independence score (as in the objective term) on a held-out source domains set. Results highlight that the proposed TCRI achieves the best worst-case domain performance.}
    \renewcommand{\arraystretch}{1.1}
    \input{tables/pacs_v3}
\end{table}

\begin{table}[h]
    \centering
    \caption{Worst-Case--PACS. (Oracle) Domain accuracies -- model selection via held-out accuracy on target domain set (i.e., unobservable in our problem setting). Oracle setting further highlights the soundness of the regularization approach, showing that it does not reject good models.}
    \renewcommand{\arraystretch}{1.1}
    \input{tables/pacs_v3_oracle}
\end{table}

\subsection{Worst-Case--VLCS}
\textbf{Data Description:} $X$: images, $Y$: animate (bird, dog, person) vs. inanimate (car, chair).

Domains: \{\{SUN09, LabelMe\}, \{LabelMe, SUN09\}, \{VOC2007\}\} \citep{Fang2013UnbiasedML}. The VOC2007 is the same as in the original dataset. In the \{SUN09, LabelMe\} domain, animate samples are selected from the SUN09 domain, while inanimate samples are selected from the LabelMe domain. Conversely, in the \{LabelMe, SUN09\} domain, animate samples are selected from the LabelMe domain, while inanimate samples are selected from the SUN09 domains. The correlation between spurious domain-specific information and the label is flipped between the first two domains and, therefore will not generalize.

\begin{table}[h]
    \centering
    \caption{Worst-Case--VLCS. Domain accuracies and all-domain statistics. ci indicates model selection via the lowest conditional independence score (as in the objective term) on a held-out source domains set. Results highlight that the proposed TCRI achieves the best worst-case domain performance.}
    \renewcommand{\arraystretch}{1.1}
    \input{tables/vlcs_v3}
\end{table}

\begin{table}[h]
    \centering
    \caption{Worst-Case--VLCS. (Oracle) Domain accuracies -- model selection via held-out accuracy on target domain set (i.e., unobservable in our problem setting). Oracle setting further highlights the soundness of the regularization approach, showing that it does not reject good models.}
    \renewcommand{\arraystretch}{1.1}
    \input{tables/vlcs_v3_oracle}
\end{table}

\newpage
\section{Theoretical Results} \label{sec:proof}
\begin{lemma}[Insufficiency of Causal Subsets for domain generalization] 
 Conditioning on a subset of causal variables (invariant mechanisms) does not imply domain generalization (definition \ref{def:dom-gen}). \[\mathcal{Z} \subset \mathcal{Z}_c \notimplies \mu_{e_i}(y \,|\, Z=z) = \mu_{e_j}(y \,|\, Z =z) \forall e_i \ne e_j,\, z \in \mathcal{Z}\]
where $\mathcal{Z}_c$ is the causal feature space.
\begin{proof}
We provide a simple counterexample. Suppose we have the following generative process $d_{e_i,1} \rightarrow z_1 \rightarrow y \leftarrow  z_2 \leftarrow d_{e_i,2}$ with $z_1, z_2 \in \mathbb{R}$, $d_{e_i,1}, d_{e_i,2} \in \mathbb{R}_{\ge 0}$, $r \in \mathbb{R}_{> 0}$, and
\begin{equation*}
    y = 
    \begin{cases}
     0 \text{ if } z_1^2 + z_2^2 <= r\\
     1 \text{ if } z_1^2 + z_2^2 > r.
    \end{cases}
\end{equation*}
Suppose we observe domains $\{e_i: i=1,\ldots,E_{tr}\}$, where $z_1 \sim \text{Uniform}(-r+d_{e_i,1}, r+d_{e_i,1})$ and $z_2 \sim \text{Uniform}(-r+d_{e_i,2}, r+d_{e_i,2})$ for domain $e_i$ and $d_{e_i}$ is a domain-specific quantity. Now, suppose we condition on $z_1$. The optimal predictor is \[\mu_{e_i}(y=1 \,|\, z_1) = \frac{2\sqrt{r^2 - z_1^2} - d_{e_i,2}}{2r}.\]

In this case, $z_1$, a causal subset, does not yield domain-general representation since its optimal predictor depends on $d_{e_i}$, which changes across domains.
\end{proof}
\end{lemma}

\begin{figure*}[t!] 
    \centering
    \resizebox{0.75\textwidth}{!}{\input{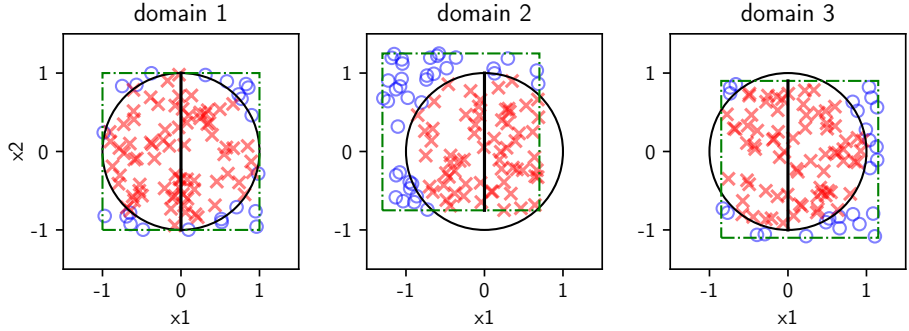}}
    \caption{Visualization of Lemma \ref{lem:subset}, where $x_1 = z_1$ and $x_2 = z_2$. The large circle is the true decision boundary, where $\textcolor{red}{\times}:\, y=0,\, \textcolor{blue}{\circ}:\, y=1$. The dotted square indicates the covariate distribution (Uniform), which differs across domains. The length of the solid line indicates the $\mu(y=0 \,|\, z_1=0)$. Clearly the length of each line changes across domains; hence the causal subset $z_1$ does not yield a domain-general predictor.}
    \label{fig:lemma1}
\end{figure*}

\begin{lemma} 
A representation $\Phi$ that maps to a strictly causal subset can be Domain-invariant. 
\begin{proof}
This proof follows a similar argument as Lemma \ref{lem:subset}. We replace the mechanism for $z_1$ with $z_1 \sim \text{Uniform}(d_{e_i,1}\cdot -r, d_{e_i, 1}\cdot r)$, i.e., the domain is scaled along $z_1$. Suppose $d_{e_i,2}=0$ for all training domains $\{e_i: i=1,\ldots,E_{tr}\}$ and $\Phi(z_1, z_2) = z_1$. There is still a covariate distribution shift due to varying values (scaling) of $d_{e_i,1}$. Then $\Phi$ is domain invariant on the training domain but will no longer be domain-general with respect to any domain $e_j$ where $d_{e_j,2} > 0$.
\end{proof}
\end{lemma}

\newpage
\begin{lemma} 
(Sufficiency of TCRI for Causal Aggregation). Recall, $X,\, Z_c,\, Z_e,\, Y$ from Figure \ref{fig:graph}. Let $Z_c, Z_e$ be direct causes and direct effects of $Y$, respectively, and recall that $X$ is a function of $Z_c$ and $Z_e$. If the two representations induced by feature extractors $\Phi, \Psi$ satisfy TCRI, then wlog $I(\Phi(X);  Y) \ge I(Z_c; Y)$.

\begin{proof} ~\\
\begin{enumerate}[label=(\roman*)]
    \item First we define $K$ $Z_c^i$'s to be random variables with non-zero mutual information with $Z_c$ marginally and conditioned on $Y$: $I(Z_c^i; Z_c) > 0$, $I(Z_c^i; Z_c \,|\, Y) > 0$.
    \item We have from (i.) that for any pair $Z_c^i, Z_c^j \in \{Z_c^1, \ldots, Z_c^K\}$, $I(Z_c^i; Z_c^j \,|\, Y) \ge 0$, since neither can be made conditionally independent of $Z_c$ given $Y$.
    \item Given the total-chain-information criterion, we have that there exist a set of \textit{K} ${Z{_c}^{i}}'s$ across $\Phi(X), \Psi(X)$ s.t. $I(Z_c^1, \ldots, Z_c^K; Y) \ge I(Z_c; Y)$ for some $K$.
    \item Combining (ii) and (iii), we have that all ${Z_{c}^{i}}'s$ are aggregated in one of the two representations, say $\Phi$, since for any $Z_c^k$ that satisfies (i.), (ii.) $\implies Z_c^k \in \Phi(X) \forall k$, and therefore (iii.) $\implies I(\Phi(X);  Y) \ge I(Z_c; Y)$.
\end{enumerate}
\end{proof}
\end{lemma}

\begin{remark}
(Revisiting Lemma \ref{lem:subset}'s counterexample) Given two representations $\Phi, \Psi$ that satisfy TCRI, $\Phi$ necessarily captures $z_1, z_2$. By definition, $\Phi, \Psi$ must capture all of the information in $z_1, z_2$ about $y$, and we know from the graph that they are conditionally dependent given $y$, i.e.,  $z_1, z_2$ are common causes of $y$ (colliders), so conditioning on $y$ renders the marginally independent variables dependent. So, $z_1$ and $z_2$ must be captured the same feature extractor.
\end{remark}

%% file: tables/cmnist_oracle.tex
\begin{tabular}{l|c|c|c} 
\textbf{Algorithm} & \textbf{+90\%} & \textbf{+80\%} & \textbf{-90\%} \\ 
\hline
ERM                  & 61.9       & 66.3       & 26.5      \\ 
IRM                  & 73.0       & 72.2       & 51.0      \\ 
GroupDRO             & 64.8       & 68.0       & 26.0      \\ 
MLDG                 & 68.8       & 72.7       & 28.6      \\ 
ARM                  & 81.6       & 73.4       & 24.2      \\ 
VREx                 & 70.2       & 70.8       & 49.7      \\ 
\hline                                                                          
TCRI (cov)           & 61.6       & 66.4       & 53.0      \\ 
TCRI (HSIC)          & 68.2       & 67.7       & 56.4      \\ 
\end{tabular}

%% file: tables/cmnist3_oracle.tex
\begin{tabular}{l|c|c|c|c} 
\textbf{Algorithm}   & \textbf{+90\%}       & \textbf{+80\%}       & \textbf{+70\%}       & \textbf{-90\%}       \\ 
\hline
ERM                  & 72.8 $\pm$ 0.3       & 74.7 $\pm$ 0.3       & 73.3 $\pm$ 0.1       & 16.3 $\pm$ 1.5      \\ 
IRM                  & 49.0 $\pm$ 0.1       & 54.2 $\pm$ 2.0       & 50.3 $\pm$ 0.3       & 43.8 $\pm$ 2.8      \\ 
GroupDRO             & 71.0 $\pm$ 0.6       & 72.2 $\pm$ 0.3       & 70.7 $\pm$ 0.9       & 36.4 $\pm$ 4.2      \\ 
MLDG                 & 72.8 $\pm$ 0.9       & 74.8 $\pm$ 0.3       & 72.9 $\pm$ 0.3       & 13.6 $\pm$ 0.7      \\ 
ARM                  & 74.7 $\pm$ 0.4       & 74.1 $\pm$ 0.2       & 73.1 $\pm$ 0.4       & 14.0 $\pm$ 1.5      \\ 
VREx                 & 74.1 $\pm$ 1.3       & 72.6 $\pm$ 0.5       & 72.1 $\pm$ 0.5       & 19.5 $\pm$ 5.5      \\ 
\hline 
TCRI  (cov)          & 68.5 $\pm$ 4.4       & 66.4 $\pm$ 6.5       & 67.8 $\pm$ 2.9       & 53.6 $\pm$ 2.3      \\ 
TCRI (HSIC)          & 72.1 $\pm$ 1.5       & 73.6 $\pm$ 0.4       & 72.6 $\pm$ 0.4       & 49.9 $\pm$ 0.3      \\ 
\end{tabular}

%% file: tables/cmnist8_oracle.tex
\begin{tabular}{l|c|c|c|c} 
\textbf{Algorithm}   & \textbf{+90\%}       & \textbf{+80\%}       & \textbf{-80\%}       & \textbf{-90\%}       \\ 
\hline
ERM                  & 58.4 $\pm$ 1.3       & 67.0 $\pm$ 0.5       & 64.2 $\pm$ 2.0       & 52.6 $\pm$ 3.2      \\ 
IRM                  & 56.7 $\pm$ 3.3       & 56.6 $\pm$ 2.8       & 51.6 $\pm$ 0.7       & 51.7 $\pm$ 0.7      \\ 
GroupDRO             & 69.7 $\pm$ 0.8       & 71.7 $\pm$ 0.3       & 72.0 $\pm$ 0.2       & 71.4 $\pm$ 1.9      \\ 
MLDG                 & 60.6 $\pm$ 0.3       & 64.6 $\pm$ 1.0       & 66.7 $\pm$ 0.5       & 55.6 $\pm$ 2.4      \\ 
ARM                  & 67.5 $\pm$ 0.4       & 65.5 $\pm$ 1.6       & 66.7 $\pm$ 0.6       & 64.7 $\pm$ 1.1      \\ 
VREx                 & 67.4 $\pm$ 1.9       & 70.4 $\pm$ 0.1       & 71.2 $\pm$ 0.2       & 59.4 $\pm$ 4.3      \\ 
\hline
TCRI (cov)           & 67.6 $\pm$ 0.8       & 64.0 $\pm$ 5.4       & 63.0 $\pm$ 5.5       & 61.5 $\pm$ 4.6      \\ 
TCRI (HSIC)          & 62.2 $\pm$ 4.4       & 70.0 $\pm$ 1.3       & 67.9 $\pm$ 1.4       & 65.4 $\pm$ 2.8      \\ 
\end{tabular}

%% file: tables/pacs_v3_oracle.tex
\begin{tabular}{l|c|c|c} 
\textbf{Algorithm} & \textbf{C x A} & \textbf{A x C} & \textbf{P} \\ 
\hline
ERM                  & 38.4 $\pm$ 1.4       & 43.4 $\pm$ 1.9       & 95.9 $\pm$ 0.6      \\ 
IRM                  & 62.8 $\pm$ 0.1       & 53.9 $\pm$ 6.6       & 85.8 $\pm$ 8.2      \\ 
GroupDRO             & 40.0 $\pm$ 1.6       & 49.7 $\pm$ 2.9       & 95.7 $\pm$ 0.6      \\ 
MLDG                 & 44.6 $\pm$ 5.5       & 40.6 $\pm$ 0.5       & 96.2 $\pm$ 0.5      \\ 
ARM                  & 44.2 $\pm$ 2.6       & 45.5 $\pm$ 2.8       & 94.3 $\pm$ 0.7      \\ 
VREx                 & 55.8 $\pm$ 5.5       & 38.7 $\pm$ 0.9       & 93.8 $\pm$ 0.8      \\ 
 
\hline 
TCRI (cov)           & 62.8 $\pm$ 0.1       & 62.3 $\pm$ 0.2       & 65.0 $\pm$ 0.4      \\ 
TCRI (HSIC)          & 64.0 $\pm$ 0.7       & 62.3 $\pm$ 0.2       & 82.4 $\pm$ 5.7      \\ 
\end{tabular}

%% file: tables/vlcs_v3_oracle.tex
\begin{tabular}{l|c|c|c} 
\textbf{Algorithm} & \textbf{S x L} & \textbf{L x S} & \textbf{V} \\ 
\hline
ERM                  & 41.1 $\pm$ 2.5       & 50.8 $\pm$ 2.7       & 82.9 $\pm$ 0.9      \\ 
IRM                  & 49.6 $\pm$ 0.4       & 53.2 $\pm$ 1.2       & 65.5 $\pm$ 0.9      \\ 
GroupDRO             & 45.3 $\pm$ 0.6       & 52.5 $\pm$ 1.1       & 78.1 $\pm$ 4.6      \\ 
MLDG                 & 40.2 $\pm$ 2.7       & 49.2 $\pm$ 0.8       & 75.6 $\pm$ 4.6      \\ 
ARM                  & 42.3 $\pm$ 0.7       & 47.4 $\pm$ 1.9       & 80.5 $\pm$ 2.5      \\ 
VREx                 & 47.1 $\pm$ 1.0       & 51.6 $\pm$ 0.4       & 78.3 $\pm$ 3.3      \\ 
 
\hline 
TCRI (cov)           & 49.6 $\pm$ 0.4       & 50.4 $\pm$ 0.1       & 73.9 $\pm$ 3.0      \\ 
TCRI (HSIC)          & 54.5 $\pm$ 2.0       & 52.9 $\pm$ 1.9       & 79.0 $\pm$ 4.6      \\ 
\end{tabular}

%% file: main.bbl
\begin{thebibliography}{29}
\providecommand{\natexlab}[1]{#1}
\providecommand{\url}[1]{\texttt{#1}}
\expandafter\ifx\csname urlstyle\endcsname\relax
  \providecommand{\doi}[1]{doi: #1}\else
  \providecommand{\doi}{doi: \begingroup \urlstyle{rm}\Url}\fi

\bibitem[Amodei et~al.(2016)Amodei, Olah, Steinhardt, Christiano, Schulman, and
  Man{\'e}]{Amodei2016ConcretePI}
Dario Amodei, Christopher Olah, Jacob Steinhardt, Paul~Francis Christiano, John
  Schulman, and Dandelion Man{\'e}.
\newblock Concrete problems in ai safety.
\newblock \emph{ArXiv}, abs/1606.06565, 2016.

\bibitem[Arjovsky et~al.(2019)Arjovsky, Bottou, Gulrajani, and
  Lopez-Paz]{Arjovsky2019InvariantRM}
Mart{\'i}n Arjovsky, L.~Bottou, Ishaan Gulrajani, and David Lopez-Paz.
\newblock Invariant risk minimization.
\newblock \emph{ArXiv}, abs/1907.02893, 2019.

\bibitem[Ben-David et~al.(2009)Ben-David, Blitzer, Crammer, Kulesza, Pereira,
  and Vaughan]{BenDavid2009ATO}
Shai Ben-David, John Blitzer, K.~Crammer, A.~Kulesza, Fernando~C Pereira, and
  Jennifer~Wortman Vaughan.
\newblock A theory of learning from different domains.
\newblock \emph{Machine Learning}, 79:\penalty0 151--175, 2009.

\bibitem[Courty et~al.(2017)Courty, Flamary, Habrard, and
  Rakotomamonjy]{courty2017joint}
Nicolas Courty, R{\'e}mi Flamary, Amaury Habrard, and Alain Rakotomamonjy.
\newblock Joint distribution optimal transportation for domain adaptation.
\newblock \emph{Advances in Neural Information Processing Systems}, 30, 2017.

\bibitem[Fang et~al.(2013)Fang, Xu, and Rockmore]{Fang2013UnbiasedML}
Chen Fang, Ye~Xu, and Daniel~N. Rockmore.
\newblock Unbiased metric learning: On the utilization of multiple datasets and
  web images for softening bias.
\newblock \emph{2013 IEEE International Conference on Computer Vision}, pages
  1657--1664, 2013.

\bibitem[Filos et~al.(2020)Filos, Tigas, McAllister, Rhinehart, Levine, and
  Gal]{Filos2020CanAV}
Angelos Filos, Panagiotis Tigas, Rowan McAllister, Nicholas Rhinehart, Sergey
  Levine, and Yarin Gal.
\newblock Can autonomous vehicles identify, recover from, and adapt to
  distribution shifts?
\newblock In \emph{ICML}, 2020.

\bibitem[Ganin et~al.(2016)Ganin, Ustinova, Ajakan, Germain, Larochelle,
  Laviolette, Marchand, and Lempitsky]{ganin2016domain}
Yaroslav Ganin, Evgeniya Ustinova, Hana Ajakan, Pascal Germain, Hugo
  Larochelle, Fran{\c{c}}ois Laviolette, Mario Marchand, and Victor Lempitsky.
\newblock Domain-adversarial training of neural networks.
\newblock \emph{The journal of machine learning research}, 17\penalty0
  (1):\penalty0 2096--2030, 2016.

\bibitem[Glover(2011)]{Glover2011OverviewOF}
Gary~H. Glover.
\newblock Overview of functional magnetic resonance imaging.
\newblock \emph{Neurosurgery clinics of North America}, 22 2:\penalty0 133--9,
  vii, 2011.

\bibitem[Gretton et~al.(2007)Gretton, Fukumizu, Teo, Song, Sch{\"o}lkopf, and
  Smola]{Gretton2007AKS}
A.~Gretton, K.~Fukumizu, C.~Teo, Le~Song, B.~Sch{\"o}lkopf, and Alex Smola.
\newblock A kernel statistical test of independence.
\newblock In \emph{NIPS}, 2007.

\bibitem[Gulrajani and Lopez{-}Paz(2020)]{Gul2020LostDG}
Ishaan Gulrajani and David Lopez{-}Paz.
\newblock In search of lost domain generalization.
\newblock \emph{CoRR}, abs/2007.01434, 2020.
\newblock URL \url{https://arxiv.org/abs/2007.01434}.

\bibitem[Hardt et~al.(2016)Hardt, Price, and Srebro]{hardt2016equality}
Moritz Hardt, Eric Price, and Nati Srebro.
\newblock Equality of opportunity in supervised learning.
\newblock \emph{Advances in neural information processing systems}, 29, 2016.

\bibitem[He et~al.(2016)He, Zhang, Ren, and Sun]{he2016deep}
Kaiming He, Xiangyu Zhang, Shaoqing Ren, and Jian Sun.
\newblock Deep residual learning for image recognition.
\newblock In \emph{Proceedings of the IEEE conference on computer vision and
  pattern recognition}, pages 770--778, 2016.

\bibitem[Kpotufe and Martinet(2018)]{pmlr-v75-kpotufe18a}
Samory Kpotufe and Guillaume Martinet.
\newblock Marginal singularity, and the benefits of labels in covariate-shift.
\newblock In Sébastien Bubeck, Vianney Perchet, and Philippe Rigollet,
  editors, \emph{Proceedings of the 31st Conference On Learning Theory},
  volume~75 of \emph{Proceedings of Machine Learning Research}, pages
  1882--1886. PMLR, 06--09 Jul 2018.
\newblock URL \url{https://proceedings.mlr.press/v75/kpotufe18a.html}.

\bibitem[Krueger et~al.(2021)Krueger, Caballero, Jacobsen, Zhang, Binas, Priol,
  and Courville]{Krueger2021OutofDistributionGV}
David Krueger, Ethan Caballero, J.~Jacobsen, A.~Zhang, Jonathan Binas,
  R{\'e}mi~Le Priol, and Aaron~C. Courville.
\newblock Out-of-distribution generalization via risk extrapolation (rex).
\newblock In \emph{ICML}, 2021.

\bibitem[Li et~al.(2017)Li, Yang, Song, and Hospedales]{li2017deeper}
Da~Li, Yongxin Yang, Yi-Zhe Song, and Timothy~M Hospedales.
\newblock Deeper, broader and artier domain generalization.
\newblock In \emph{Proceedings of the IEEE international conference on computer
  vision}, pages 5542--5550, 2017.

\bibitem[Li et~al.(2018)Li, Yang, Song, and Hospedales]{li2018learning}
Da~Li, Yongxin Yang, Yi-Zhe Song, and Timothy~M Hospedales.
\newblock Learning to generalize: Meta-learning for domain generalization.
\newblock In \emph{Thirty-Second AAAI Conference on Artificial Intelligence},
  2018.

\bibitem[Long et~al.(2016)Long, Zhu, Wang, and Jordan]{long2016unsupervised}
Mingsheng Long, Han Zhu, Jianmin Wang, and Michael~I Jordan.
\newblock Unsupervised domain adaptation with residual transfer networks.
\newblock \emph{Advances in neural information processing systems}, 29, 2016.

\bibitem[Pearl(2010)]{Pearl2010CausalI}
J.~Pearl.
\newblock Causal inference.
\newblock In \emph{NIPS Causality: Objectives and Assessment}, 2010.

\bibitem[Peters et~al.(2016)Peters, B{\"u}hlmann, and
  Meinshausen]{Peters2015CausalIU}
Jonas Peters, Peter B{\"u}hlmann, and Nicolai Meinshausen.
\newblock Causal inference by using invariant prediction: identification and
  confidence intervals.
\newblock \emph{Journal of the Royal Statistical Society. Series B (Statistical
  Methodology)}, pages 947--1012, 2016.

\bibitem[Rosenfeld et~al.(2020)Rosenfeld, Ravikumar, and
  Risteski]{Rosenfeld2021TheRO}
Elan Rosenfeld, Pradeep Ravikumar, and Andrej Risteski.
\newblock The risks of invariant risk minimization.
\newblock \emph{arXiv preprint arXiv:2010.05761}, 2020.

\bibitem[Rosenfeld et~al.(2022)Rosenfeld, Ravikumar, and
  Risteski]{Rosenfeld2021AnOL}
Elan Rosenfeld, Pradeep Ravikumar, and Andrej Risteski.
\newblock An online learning approach to interpolation and extrapolation in
  domain generalization.
\newblock In \emph{International Conference on Artificial Intelligence and
  Statistics}, pages 2641--2657. PMLR, 2022.

\bibitem[Sagawa et~al.(2019)Sagawa, Koh, Hashimoto, and
  Liang]{sagawa2019distributionally}
Shiori Sagawa, Pang~Wei Koh, Tatsunori~B Hashimoto, and Percy Liang.
\newblock Distributionally robust neural networks for group shifts: On the
  importance of regularization for worst-case generalization.
\newblock \emph{arXiv preprint arXiv:1911.08731}, 2019.

\bibitem[Schrouff et~al.(2022)Schrouff, Harris, Koyejo, Alabdulmohsin,
  Schnider, Opsahl-Ong, Brown, Roy, Mincu, Chen,
  et~al.]{schrouff2022maintaining}
Jessica Schrouff, Natalie Harris, Oluwasanmi Koyejo, Ibrahim Alabdulmohsin, Eva
  Schnider, Krista Opsahl-Ong, Alex Brown, Subhrajit Roy, Diana Mincu,
  Christina Chen, et~al.
\newblock Maintaining fairness across distribution shift: do we have viable
  solutions for real-world applications?
\newblock \emph{arXiv preprint arXiv:2202.01034}, 2022.

\bibitem[Shen et~al.(2021)Shen, Liu, He, Zhang, Xu, Yu, and
  Cui]{shen2021towards}
Zheyan Shen, Jiashuo Liu, Yue He, Xingxuan Zhang, Renzhe Xu, Han Yu, and Peng
  Cui.
\newblock Towards out-of-distribution generalization: A survey.
\newblock \emph{arXiv preprint arXiv:2108.13624}, 2021.

\bibitem[Singh et~al.(2021)Singh, Singh, Mhasawade, and
  Chunara]{singh2021fairness}
Harvineet Singh, Rina Singh, Vishwali Mhasawade, and Rumi Chunara.
\newblock Fairness violations and mitigation under covariate shift.
\newblock In \emph{Proceedings of the 2021 ACM Conference on Fairness,
  Accountability, and Transparency}, pages 3--13, 2021.

\bibitem[Vapnik(1991)]{NIPS1991_ff4d5fbb}
Vladimir Vapnik.
\newblock Principles of risk minimization for learning theory.
\newblock In \emph{NIPS}, volume~91, pages 831--840, 1991.

\bibitem[Wang et~al.(2022)Wang, Si, Li, and Zhao]{wang2022provable}
Haoxiang Wang, Haozhe Si, Bo~Li, and Han Zhao.
\newblock Provable domain generalization via invariant-feature subspace
  recovery.
\newblock In \emph{ICML}, 2022.

\bibitem[Zhang et~al.(2021)Zhang, Marklund, Dhawan, Gupta, Levine, and
  Finn]{zhang2021adaptive}
Marvin Zhang, Henrik Marklund, Nikita Dhawan, Abhishek Gupta, Sergey Levine,
  and Chelsea Finn.
\newblock Adaptive risk minimization: Learning to adapt to domain shift.
\newblock \emph{Advances in Neural Information Processing Systems}, 34, 2021.

\bibitem[Zhao et~al.(2019)Zhao, des Combes, Zhang, and Gordon]{Zhao2019OnLI}
H.~Zhao, R{\'e}mi~Tachet des Combes, Kun Zhang, and Geoffrey~J. Gordon.
\newblock On learning invariant representations for domain adaptation.
\newblock In \emph{ICML}, 2019.

\end{thebibliography}
